# Bridging the Gap with Retrieval-Augmented Generation: Making Prosthetic Device User Manuals Available in Marginalised Languages

Ikechukwu Ogbonna[1,*], Lesley Davidson[2], Soumya Banerjee[3], Abhishek Dasgupta[4], Laurence Kenney[2], and Vikranth Harthikote Nagaraja[2,#]

[1]Department of Biomechanical Engineering, Faculty of Science and Technology, University of Twente, Enschede, Netherlands;
[2]Centre for Human Movement and Rehabilitation, School of Health and Society, University of Salford, Salford, M6 6PU, UK;
[3]Department of Computer Science & Technology, University of Cambridge, 15 J.J. Thomson Ave., Cambridge, CB3 0FD, UK;
[4]Doctoral Training Centre, University of Oxford, 1–4 Keble Road, Oxford, OX1 3NP, UK.
[*]i.i.ogbonna@student.utwente.nl | [#]v.harthikotenagaraja@salford.ac.uk

## ABSTRACT

Millions of people in African countries face barriers to accessing healthcare due to language and literacy gaps. This research tackles this challenge by transforming complex medical documents—in this case, prosthetic device user manuals—into accessible formats for underserved populations. This case study in cross-cultural translation is particularly pertinent/relevant for communities that receive donated prosthetic devices but may not receive the accompanying user documentation. Or, if available online, may only be available in formats (e.g., language and readability) that are inaccessible to local populations (e.g., English-language, high resource settings/cultural context). The approach is demonstrated using the widely spoken Pidgin dialect, but our open-source framework has been designed to enable rapid and easy extension to other languages/dialects. This work presents an AI-powered framework designed to process and translate complex medical documents, e.g., user manuals for prosthetic devices, into marginalised languages. The system enables users—such as healthcare workers or patients—to upload English-language medical equipment manuals, pose questions in their native language, and receive accurate, localised answers in real time. Technically, the system integrates a Retrieval-Augmented Generation (RAG) pipeline for processing and semantic understanding of the uploaded manuals. It then employs advanced Natural Language Processing (NLP) models for generative question-answering and multilingual translation. Beyond simple translation, it ensures accessibility to device instructions, treatment protocols, and safety information, empowering patients and clinicians to make informed healthcare decisions. This framework supports integrating additional languages, making it adaptable to a wide range of global health challenges, including public health campaigns and disaster relief, where accurate communication in native languages can save lives. With far-reaching implications, this research serves as an interim solution for health organisations/providers in such communities and a call to action for policymakers and governments of the Global South to reduce inequities in accessing critical medical information. Overall, it offers a vision of healthcare that empowers marginalised communities, fosters trust, and ensures no one is left behind due to language or literacy barriers.

## 1 Introduction

Prosthetic manufacturers provide support for their products, and the relevant details are typically outlined within the product documentation. User manuals are crucial since they provide end users with critical information regarding safety, device limitations, and appropriate use. Such guidelines are intended to influence user behaviour(s) with their prosthetic limb. However, in African countries, it is common practice to accept second-hand assistive technologies like prosthetic devices donated by organisations located in high-resource settings [1]. Although some prosthetic companies sell devices in various African countries, user manuals for these markets are rarely, if ever, provided in the variety of formats and languages needed to suit most of the local populations. This could potentially be attributed to a lax regulatory framework and/or user manuals being an optional requirement in the recipient's region. Prosthetic user manuals provide helpful training information to users, with implications on user outcomes [2], [3]. Studies [4], [5] have highlighted that a lack of access to appropriate manuals may lead to an increase in the likelihood of frustration, improper use, and even harm when using the devices. Further, it may contribute to device underutilisation, and hence, waste [4], [5]. This work explores the use of generative artificial intelligence (AI) to process complex medical manuals and transform them into accessible, localised languages understood by people of underserved populations or marginalised languages. Unfortunately, most of the world's languages have been marginalised due to the overrepresentation of the highest-resource languages (e.g., English) through online activity on the internet. Consequently, due to such linguistic imbalances, AI models have been found to underperform for speakers of low-resource languages [6] – [8], eventually leading to epistemic injustice [9], warranting much-needed participatory efforts in this direction. The study aims to investigate the applicability of open-source large language models (LLMs) to develop an adaptable framework capable of understanding, processing, and presenting information from user manuals in clear, localised languages understandable to individuals with varying literacy levels. It also aims to promote inclusivity and accessibility by designing a framework that can be replicated in other languages, particularly focusing on low-resource and marginalised languages.

## 2 Methods

Here, we took prosthetic user manuals for use in Nigeria as a case study. Nigeria is the most populous country in Africa and the sixth most populous in the world, and is home to 230 million people [10]. Despite its population, very little NLP research has been done on its over 500 indigenous languages, including the major languages spoken (yet underrepresented online)—Igbo, Yoruba, Hausa, and Pidgin [11].

Initially, we attempted to frame the problem as a text simplification problem [12], [13]. The idea was to develop an NLP pipeline that assessed the readability of texts in prosthetic user manuals written in English, after which it converted those texts into various literacy levels in Igbo and Yoruba. However, this approach was limited by a lack of available dictionaries and datasets required to train text simplification models on the Igbo and Yoruba languages. Another limitation was the lack of readability evaluation metrics for Igbo and Yoruba, and without evaluation metrics, the crucial question *"Do you simplify the texts in English and then translate to Igbo or do you translate to Igbo and then simplify the Igbo text?"* could not be answered. Hence, raising uncertainties about the most efficient approach to solve the initial problem. The above-mentioned limitations raised the question, *"How do we solve the initial problem without reinventing the wheel for every marginalised language?"*.

The Question Answering approach was our answer. Question Answering (QA) is an NLP task where models are used to extract or generate precise answers to questions asked by humans in natural language based on either a pre-structured database or a collection of natural language documents. It enables efficient information retrieval, comprehension, and decision-making [14] – [17]. The idea was to have a system in which users can upload their user manuals (in English), then ask questions in their native language and get back the answer in their native language. The target language upon which the system was developed also changed from Igbo and Yoruba to Pidgin because it and its variants are spoken not only in Nigeria but across West and Central Africa, making this solution potentially more widely useful [18]. This idea relies on the rapid availability of a translation model for all languages as part of initiatives to bring national development and bridge the AI gap being spearheaded by the United Nations (UN) and private organisations like Google and Meta. The idea then assumes that translation models are trained with dictionaries that are understandable to all native speakers of that language, irrespective of educational qualification. Our open-source model and system architecture are accessible on GitHub: https://github.com/Iykay/User-Manual-LLM.

## 3 Results and Conclusion

The result of the development process is an amalgamation of open-source models. Our scalable framework uses English-to-Pidgin as the case study. However, the underlying architecture could be deployed with other languages provided translation models are available, and the source user manual is in any of the 20 typical high-resource languages [19]. It is a plug-and-play situation; no additional support is required.

Nonetheless, further work is required before rolling out this product in a real-world setting including: (i) optimisation for image and table handling such that results of queries also return associated images and tables; (ii) language expansion involving the incorporation of more languages than Pidgin; (iii) data expansion involving collation, indexing and storage of prosthetic user manuals in a database so that users do not need to find a PDF file to upload to the system; (iv) reduction in processing time; (v) user interface testing and improvement; (vi) back translation, checking of equivalences, clarity and relevance, usability testing; and (vii) ethical, responsible, and regulatory alignment/guardrails [20], [21] to ensure that the system always provides information in line with the safety of the user. Perhaps, including an expert-in-the-loop might help enhance the trust and reliability of generated texts and provide feedback to improve model performance.

In conclusion, by developing this framework, we aim to facilitate ongoing efforts to improve health literacy, global health equity, and health outcomes in line with the UN Sustainable Development Goals (including 'SDG 3: Good Health and Well-Being'). Additionally, this work aligns with the UN's goals of building a more inclusive digital future through greater representation and inclusion of diverse languages [11], [22]. We hope to stir leaders of industries, policymakers, and governments in countries where these regulations are missing to create policies that protect the rights of their members to access health and health-related information in an appropriate language and format. Finally, we hope to inspire scientists, inventors, and enthusiasts in the Global South to abandon hedonistic or wasteful attempts at reinventing the wheel by leveraging open-source resources. We hope to inspire them to become change agents to reimagine the problems they face daily and consider how they can adapt already existing tools and frameworks to address problems that exist in their daily lives.